\documentclass[12pt]{amsart}

\usepackage{amsfonts,amssymb,latexsym}
\usepackage{amsthm}
\usepackage{color,cancel}

\usepackage{url}

\usepackage{float}
\usepackage{graphicx}

\begin{document}
\title[Facial reconstruction method using latent root regression]{ Craniofacial reconstruction as a prediction problem using a latent root regression model}

\author[Berar et al.]{Maxime Berar
, Francoise M. Tilotta, Joan A. Glaun\`es  and Yves Rozenholc
}


\email{\newline 
Berar M. (Corresponding author) : maxime.berar@univ-rouen.fr
}

\begin{abstract}
In this paper, we present a computer-assisted method for facial reconstruction. This method provides an estimation of the facial shape associated with unidentified skeletal remains. Current computer-assisted methods using a statistical framework rely on a common set of extracted points located on the bone and soft-tissue surfaces. Most of the facial reconstruction methods then consist in predicting the position of the soft-tissue surface points, when  the positions of the bone surface points are known. We propose to use Latent Root Regression for prediction. The results obtained are then compared to those given by Principal Components Analysis linear models. In conjunction, we have evaluated the influence of the number of skull landmarks used. Anatomical skull landmarks are completed iteratively by points located upon geodesics which link these anatomical landmarks, thus enabling us to artificially increase the number of skull points. Facial points are obtained using a mesh-matching algorithm between a common reference mesh and individual soft-tissue surface meshes. The proposed  method is validated in term of accuracy, based on a leave-one-out cross-validation test applied to a homogeneous database. Accuracy measures are obtained by computing the distance between the original face surface and its reconstruction. Finally, these results are discussed referring to current computer-assisted reconstruction facial techniques.
\end{abstract}

\maketitle


\rm
\section{Introduction}
In forensic medicine, craniofacial reconstruction refers to any process that aims to approximate the morphology of the face from the shape of the skull \cite{Wilkinson2005}. It is usually considered when confronted with an unrecognisable corpse and when no other identification evidence is available. This reconstruction may hopefully provide a route to a positive identification. In recent years, computer-assisted techniques have been developed following the evolution in medical imaging and computer science. As presented in the surveys in \cite{Buzug2006,Clement2005,DeGreef2005}, new approaches are now available with reduced performance timeline and operator subjectivity.

Reproducing the stages of manual methods, the first machine-aided techniques fitted a skin surface mask to a set of interactively-placed virtual dowels which were on the digitised model of the remains \cite{Evenhouse1992,Vanezis2000,Shahrom1996}. 
Latter techniques have moved away from the manual techniques and two kind of methods can be distinguished based on the representation of bone and soft-tissue volumes. 
The first techniques aim to conserve the continuous nature of the skull and soft-tissue surfaces. 
Estimates of the face are obtained by applying space-deformations to couples of identified bone and soft tissue surfaces, called reference surfaces. 
These deformations are learned between the surface of the dry skull and the surfaces of the reference skull. 
They can be parametric (e-g B-splines) \cite{Kermi2007,Vandermeulen2006}, implicit using variational methods \cite{Mang2006,Mang2007} or volumetric \cite{Quatrehomme1997,Nelson1998}. 
Depending on the method, the final estimated face can be either the deformed face whose reference skull is the nearest to the dry skull \cite{Quatrehomme1997,Nelson1998} or a combination of all the deformed soft-tissue surfaces \cite{Vandermeulen2006,Tu2007}.

The second type of approach chooses to represent individuals using a common set of points. 
As the position of the corresponding points for all the individual can be summarized as variables in a table, the main idea is then to use statistics to decipher the relation between the skull  and the soft-tissue. 
The common set of points can either be anatomical landmarks \cite{Vanezis2000,Claes2006} or semi-landmarks located following a point correspondence procedure \cite{Berar2006,Kahler2003}. 
Semi-landmarks are defined as points that do not have anatomical meanings but that match across all the samples of a data set under a reasonable model of deformation \cite{Bookstein1997}. 
The larger the set of points, the closer this surface representation is to a real surface.

Apart from the practical constraint of the number of anatomical landmarks that an expert can define and extract, there is no justification to a chosen number of points. 
Indeed, the information given by the position of a skull anatomical landmarks is double. 
First, there is the geometric information given by the coordinates of the point.  
Then, the ``anatomic'' information is provided by the measuring of tissue thickness made at this point. 
This information is available for a limited number of points, due to the difficulties in localisation and in thickness computation. 
However, the geometric information given by the position of the point can be completed by automatic methods of semi-landmarks extraction. 
The second part of the data analysis framework consists in learning relationships between the soft-tissue variables and the bone variables. 
Facial reconstruction consists of predicting the positions of the soft-tissue points knowing the positions of the set of skull landmarks. 
In this regression problem, the skull points are entries of a regression model while the face points are the model output. 
Typically, a linear modelisation is used to fit this regression based on Principal Components Analysis (PCA) for example \cite{Tu2007,Claes2006,Berar2006}, following the work made in the statistical atlas field. 
However, several linear regression methods have been developed, some sharing the use of PCA. 
For example latent variables regression methods such as Latent Root regression \cite{Vigneau2002,Gunst76} are designed to take the presence of variable colinearity into account, namely the positions of the skull landmarks and of the face semi-landmarks.

In this paper, we propose a facial reconstruction technique using Latent Root Regression, before comparing the results given to those obtained by a PCA model. 
In conjunction, we question the number of skull landmarks necessary. 
Basing our first experimentation on skull anatomical landmarks extracted by an expert, we will iteratively add supplementary mathematical skull semi-landmarks following the point correspondence technique described in \cite{Wang01}, which relies on the geodesic paths between the landmarks to define new landmarks.

The paper is organised as follows. 
The material and method are presented in Section 2: Section 2.1 presents the material on which this study has been carried out. 
Section 2.2 and Section 2.3 focus on resolving the point correspondence problem, describing the two methods used to obtain the two subject-shared description of the bone and soft-tissue surfaces. 
Section 3 presents the statistical methods used: the building and use of a  statistical shape model and the multivariate Latent Root Regression method. 
Section 4 shows the results obtained by the different models and discusses the influence of the number of skull landmarks and of the statistical method chosen.

\section{Material and method}
\subsection{Material}
This study was performed using whole head and skull surface meshes extracted from whole head CT scanners for a project on facial reconstruction at Paris Descartes University. 
The head CT-scan database of healthy people build for this project is composed of several type of data: head CT-scans, triangulated and closed surfaces covering the skull and the face, soft-tissue measurements at predefined skull landmarks \cite{Tilotta2007,Tilotta2008arXiv}. These anatomical skull landmarks were manually located on each CT Scan according to classical methods of physical anthropology (13 midpoints and two sets of 13 lateral points). This database, and the processes performed on it, are described in detail in Tilotta et al. 2008. 

In the framework of this study, we focus on a group of 47 European female patients aged from 20 to 40 years. 
As properties such as age, gender and ancestry can be determined by anthropological examination \cite{Reichs92}, we choose to build statistical models on a homogeneous database following these criteria. 
In our case, the 85 subjects of the database were distributed according to sex and age group (20-40 years, 40-65 years) with age groups determined so as to take ageing into account. 
Women aged from 20 to 40 years correspond to the largest group of the database.
Body weight is another factor that affects facial form \cite{Starbuck07}. In our case, none of the 47 women has a Body Mass Index (BMI) superior to 30. 8 subjects are underweight (BMI $<$ 19), 34 subjects correspond to a normal weight (BMI: 19-24) and 5 subjects are overweight (BMI: 14-30). 

The entries of our database will consist of left or right halves of each surface. 
The skull and the face do not have symmetric shapes, but the relationship between these shapes do not depend on the side of the head. 
The plan minimising the distances to the anatomical midpoints has been chosen as an artificial boundary between the right and left part of the shapes. 
The next step is to extract points which correspond to the same places on the different individuals, with respect to this symmetry constraint.

\subsection{Point correspondence procedure for the bone surface}

\begin{figure}[htbp]
\begin{center}
\includegraphics[width=15.2cm]{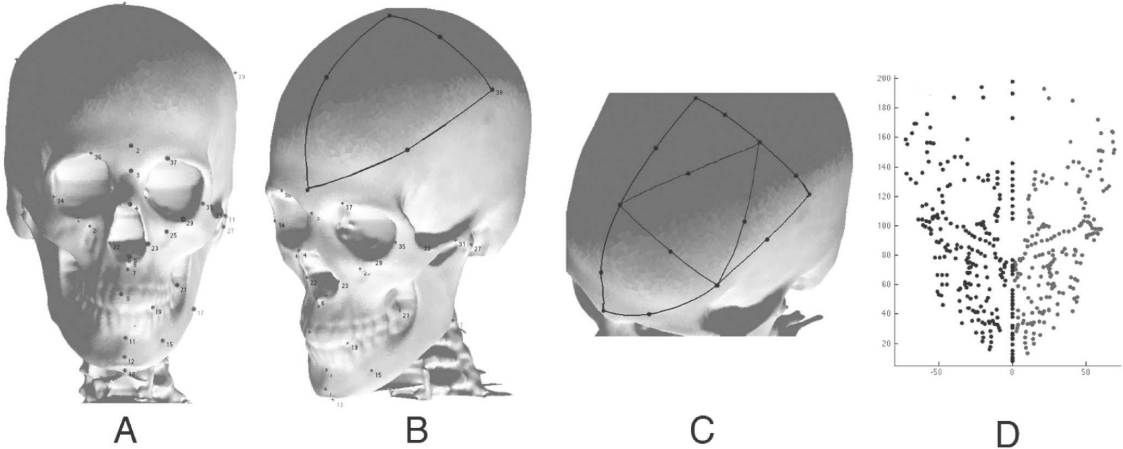}
\caption{Extracted landmarks and semi-landmarks for the skull  (\cite{Wang01})}
\label{fig:Procedure1}
\end{center}
\end{figure}

The anatomical landmarks located by the expert (Figure \ref{fig:Procedure1},A) establish a first correspondence between the skulls. 
Following the scheme presented in \cite{Wang01}, we define a set of triangular connections between these anatomical landmarks. 
For each pair of connected points, we can extract a geodesic between these points. 
Geodesics are defined to be the shortest path between points on the curved spaces of the shape surfaces (see Figure \ref{fig:Procedure1},B). 
As the shape surface between two landmarks is different from a sphere, theses geodesics are unique with high probability.  
At this step, a gross template of curbs on the surface between the landmarks is built. We then can define new landmarks as the midpoints of each geodesic.
These landmarks are used to build new geodesics as seen in Figure \ref{fig:Procedure1},C and a more dense triangulation is then derived. 
The process is iterated to form a dense geodesic triangulation with associated semi-landmarks. 
The obtained structures form meshes, which share the same structure for each individual, and implicitly solve the point correspondence problem. 
Moreover, the defined structure is symmetric: the two entries (left and right) of the database share a common substructure and set of midpoints (Figure \ref{fig:Procedure1},D). 
Due to numerical instabilities, two methods of geodesics computation on surface meshes have been used: The Surashky  algorithm \cite{Surashky2005} and the \textit{Fast Marching Algorithm} algorithm \cite{Sethian1999}  implemented by Peyre in the Geowave library. 
For two iterations of this procedure, the results show three sets of skull landmarks for each individual: the 26 original landmarks (13  midpoints), a second set composed of 55 points added by the first iteration of the procedure (10 midpoints) then completed with 164 points (24 midpoints). 
For more iteration, the point correspondence step is confronted with degenerated triangles - i.e. segments or points - in the zones where the anatomical landmarks are close to each other.

\subsection{Point correspondence procedure for the soft-tissue surface}

\begin{figure}
\begin{center}
\includegraphics[height=5.50cm]{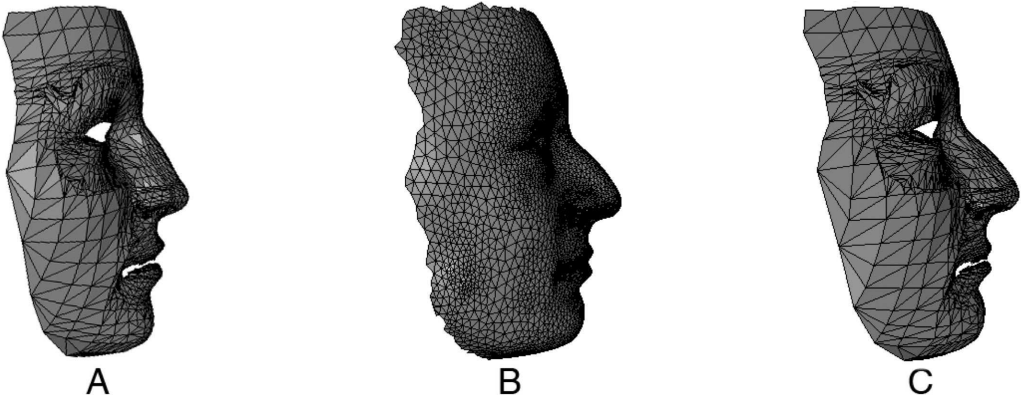}
\caption{Mesh-matching points correspondence procedure :
(A) reference mesh (B) Soft-tissue surface (C) Deformed reference mesh}
\label{fig:Procedure2}
\end{center}
\end{figure} 

For the soft-tissue surface, no landmarks are located. 
Moreover measures of tissue thickness are not provided. 
Instead of facial points analogous to the anatomical skull points, we extract for each individual a set of semi-landmarks which are neither really dense nor sparse. 
Working on the ``half''  surfaces previously defined, the point correspondence procedure registers a reference mesh (see Figure \ref{fig:Procedure2},A) on the individual soft-tissue surface (see Figure \ref{fig:Procedure2},B) resulting in a deformed reference mesh (see Figure \ref{fig:Procedure2},C). 
The mesh used as the reference mesh corresponds to the face region of the head mesh modelled by F. Pighin \cite{Pighin99}, where the density of vertices is greater in zones with high bending than in zones with low bending. 
The registration is made computing an elastic deformation between the reference mesh and soft-tissue surfaces of the database. 
The deformed meshes of each entry of the database have the same number of vertices (1741 for the mesh of a half face). 
The assumption of semi landmarks is then made: each vertex of the deformed reference meshes matches the same point for every individual. The  3D to 3D meshes matching algorithm used is a modified version of Szeliski algorithm \cite{Szeliski96}.

\begin{figure}
\begin{center}
\includegraphics[width=6.0cm]{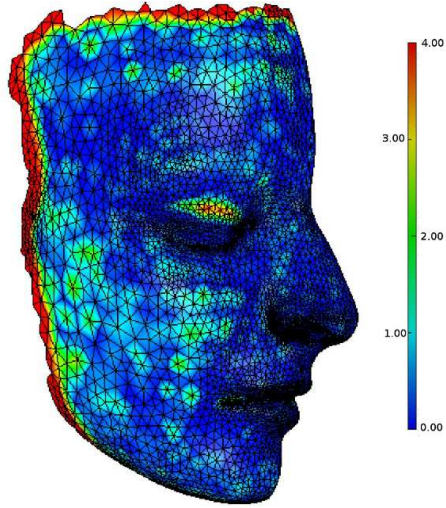}
\caption{Distance card from the vertices of the soft-tissue surface mesh to the deformed reference mesh. Right: color scheme associated to the distance values from 0.0 mm (blue) to 4.0 mm (red)}
\label{fig:BruitCorrespondence}
\end{center}
\end{figure}

This dissimilarity between the soft-tissue surface meshes and the reference meshes have consequences. 
The distances from the vertices of the deformed reference mesh to its associated soft-tissue surface mesh are null. 
But, the distances from the vertices of the soft-tissue mesh surface to the deformed reference mesh are not null, as seen in Figure \ref{fig:BruitCorrespondence}. 
The highest distances (larger than 3 mm) correspond to parts of the soft-tissue surfaces which do not have corresponding regions in the reference surface. 
Other distances correspond to regions like the forehead or the cheeks where the registration suffers from the low density of the semi-landmarks reference mesh. 
Vertices with no direct counterparts can be as far as 2 mm from the surface defined by the deformed reference mesh. 
Measures of the error introduced during this point correspondence are influenced by the large distances  generated by the lack of correspondence on the boundaries. 
The individual mean of the distances ranges from 0.54 mm to 2.66 mm. 
However, the median which correspond to the middle sample of the ordered samples, is by definition less influenced by the presence of a small number of high value outliers. 
Individual distance medians range from 0.17 mm to 0.34 mm. The mean median of distances is 0.22 mm (with standard deviation of 0.04 mm).

\section{Statistical Methods}

\subsection{Statistical Frameworks}
The variables  $\mathbf{\tilde{x}_{i}}$ are obtained from the positions of the $N$ skull points. 
The variables $\mathbf{\tilde{y}_{i}}$ are obtained from the positions of the $L$ soft-tissue points :
\begin{align}
\mathbf{\tilde x_{i} = [S_1^x\;S_1^y\;S_1^z \cdots S_N^x\;S_N^y\;S_N^y]},\\
\mathbf{\tilde y_{i} = [F_1^x\;F_1^y\;F_1^z \cdots F_L^x\;F_L^y\;F_L^z]}.
\end{align}
Two geometricaly averaged templates $ \bar{\mathbf{x}}$ and  $\bar{\mathbf{y}}$ are computed and the data centred :
\begin{align}
\mathbf{x_{i} =  \tilde{x}_{i} - \bar{x}},\\
\mathbf{y_{i} =  \tilde{y}_{i} - \bar{y}}.
\end{align}
The datatables $\mathbf{X}$ and $\mathbf{Y}$, of size $n \times 3N$ and $n \times 3L$, encompass the variables corresponding to the $\mathbf{n}$ centred samples $\mathbf{x_{i}}$ and $\mathbf{y_{i}}$ of the learning database.
Consider the matrix $\mathbf{Z = [X Y]}$ formed by merging data tables $\mathbf{X}$ and $\mathbf{Y}$ and perform Principal Component Analysis on $\mathbf{Z}$. 
The result of this PCA is a correlation-ranked set of statistically independent modes of principal variations $\mathbf{a_j}$. 
These principal modes or components are in turn vectors of 3D coordinates (of size $3(N+L)$) defined as linear combinations of points position.
These vectors capture the variations observed over all subjects in the database and are the eigenvectors of the matrix $\mathbf{^tZZ}$ associated to the eigenvalues $\lambda_{i}$ sorted such as $\lambda_{1}\geq\cdots\geq\lambda_{i}\geq \cdots\geq\lambda_{N+L}\geq0$.
\begin{align} 
\mathbf{\lambda_{i}a_{i}=\;^tZZa_{i}}.
\label{PCA}
\end{align}
Every entry $\mathbf{z_{i}}$ in the database can now be represented as a weighted linear combination of these principal components :
\begin{align}
\mathbf{z_{i}} =  \sum_{j=1}^{m}  b_{ij}\;\mathbf{a_j},
\end{align}
where $b_{ij}$ is the weight attached to sample $i$ and component $j$. 
Each eigenvector  $\mathbf{a_i}$ obtained with PCA for which $\lambda_{i}>0$ can be decomposed as the juxtaposition of two vectors 
$\mathbf{a_{i} = [v_i\;w_i]}$, with  $\mathbf{v_i} \in \Re^{3N}$ and $\mathbf{w_i} \in \Re^{3L}$. Each part $\mathbf{x_i}$ and $\mathbf{y_i}$ associated to the entry $\mathbf{z_i}$ can be represented in the model sharing the same weights $b_{ij}$ and the 3D coordinates  $\mathbf{v_i}$ and $\mathbf{w_i}$ of the vectors of the principal components  :
\begin{align}
\mathbf{x_i} = \mathbf{\tilde{x}}_i -\bar{\mathbf{x}} = \sum_{j=1}^{m} b_{ij} \mathbf{v_j},\\
\mathbf{y_i} = \mathbf{\tilde{y}}_i -\bar{\mathbf{y}} =  \sum_{j=1}^{m}  b_{ij}\mathbf{w_j}.
\end{align}

\subsection{Best Model Fit}
For facial reconstruction, we seek the best model fit of the PCA-model, i.e. the weights $b_{0j}$ of the model the nearer from the centred measured skull landmarks $\mathbf{ x_0 = \tilde{x}_0 - \bar{x} }$, which are the only points known. 
The problem is resolved finding successively each weight $b_{0j}$ and removing its contribution to the skull landmarks :
\begin{align}
&\hat{b}_{0j} = \arg\min_{b_{0j}} || \mathbf{x}_{0j-1} - b_{0j} \mathbf{v_j} ||^{2}, \\
&\mathbf{x_{0j}} = \mathbf{x_{0j-1}} - b_{0j} \mathbf{v_j},
\end{align}
with $\mathbf{ x_{00} = x_0 }$.

The solution is given by the projection of the residual centred skull $\mathbf{x_{0j}}$ landmarks upon the vector $\mathbf{ v_j}$ :
\begin{align}
\hat{b}_{0j} =  \frac{<\mathbf{x_{0j}},\mathbf{ v_j} >}{||\mathbf{v_j }||^2}.
\end{align}
and the facial reconstruction obtained is: 
\begin{align}
\mathbf{\tilde{y}_{i}} = \bar{\mathbf{y}} +  \sum_{j=1}^{m}  \hat{b}_{0j}\mathbf{w_j}.
\end{align}

\subsection{Latent Root Regression}
Latent Root Regression (LRR) is also a linear regression method. The multiresponse linear regression model for centred data is defined as :
\begin{align}
\mathbf{Y = XB + E},
\end{align}
where $\mathbf{B}$ is a $3N\times 3L$ matrix of regression coefficients and $\mathbf{E}$ is a noise matrix of size $n\times 3L$. 
The elements of the matrix  $\mathbf{E}$ are assumed to be centred with $E \mathbf{E}=\vec 0$ with covariance matrix $\Sigma$. 
Given a new set of centred skull landmarks $\mathbf{x_{0}}$, an estimate of  $\mathbf{y_{0}}$ will be: 
\begin{align}
\mathbf{\hat y_{0}= x_{0}\hat{B}}.
\end{align}
where $\hat{\mathbf{B}}$ is an estimate of $\mathbf{B}$.

LRR is similar to Partial Least Square (PLS) regression or Principal Component Regression (PCR). 
It originates from chemiometrics were a small number of predictors must predict a great number of output. 
It is then particulary adapted to the ratio between the 245 skull landmarks and the 1741 face points. Multiresponse Latent Root Regression gives an estimate of the regression matrix coefficients using Principal Components Analysis. 
Single response Latent Root Regression \cite{Gunst76} uses the same vectors $\mathbf{v_i}$ as the statistical shape model to estimate $\mathbf{B}$. 
As these vectors are not necessarily orthogonal, an iterative procedure build upon the first latent variable is needed (see Vigneau and Qannari [20] for details). 
It results in a sequence of $r$ orthogonal vectors $\mathbf{\tilde v_i}$ which enables us to compute regression coefficients $\mathbf{\hat B}$, following the formula :
\begin{align}
\mathbf{\hat B = \sum_{i=1}^r \frac{\tilde v_i\;^t \tilde v_{i}}{^t \tilde v_i\;^tXX \tilde v_i}\;^tXY}.
\end{align}
The prediction of the vector $\mathbf{\hat y_0}$ is then given by
\begin{align}
\mathbf{^t \hat y_{0}=\;^tx_{0}\sum_{i=1}^r \frac{\tilde v_i\;^t \tilde v_{i}}{^t \tilde v_i\;^tXX \tilde v_i}\;^tXY}.
\end{align}

\section{Results}

\subsection{Validation}
The validation of the proposed statistical methods for craniofacial reconstruction is obtained by a leave-one-out cross-validation procedure. 
In turn, a sample is removed from the database and used as a test case. 
The remaining samples are used to create the statistical model. 
The reconstruction procedure is applied to the test sample and a reconstruction error computed. 
Each sample is composed of two couples, left and right, of soft-tissue and bone surfaces while skull points of each couple are used as separate entries. The resulting predicted surfaces for the half of the face can then be compared with the skin surface of the test case representing the home truth.

A quantitative reconstruction error is performed by calculating the distances between every point on the reconstructed skin surface and the measured skin surface of the test case. 
Other criteria for assessing the resemblance between the reconstructed and the original face surfaces can be chosen. 
On the one hand, exterior experts can judge the resemblance between the two shapes offering a subjective evaluation of the results \cite{Claes2006, Quatrehomme2007}. 
However, this kind of evaluation needs the reconstruction process to be more complete than the prediction of the shape of surfaces, for example by  including a virtual make-up step. 
On the other hand a selection of specific measurements can be made between landmarks of the bone and tissue surfaces (for examples of such anthropological distances see \cite{Quatrehomme2007}). 
As our methodology chooses to use semi-landmarks instead of landmarks located by an expert, such a type of measurement will in our case be very dependent from the point correspondence step. 
However, the local repartition of the distance error can be computed for each semi-landmark providing distance maps which give the prediction error for each semi-landmark. 
Another point-to-surface error can be computed: the distance between the points of the skin surface of the test case and the reconstructed skin surface. 
As it has been observed during the evaluation of the mesh-matching procedure, the dissimilarity in densities of the meshes overestimates the error and the median of the distances is a good way to take these problems into account. 
Subsequently, reconstruction errors will denote the mean distance between every point on the reconstructed skin surface and the measured skin surface of the test case.\\

\subsection{Results}
The leave-one-out cross validation procedure described previously was performed on the currently available database resulting in 47 test cases. 
As the database is composed of half parts of the bone and skin surface, we can use as much as 91 components for the prediction of the soft-tissue surface. 
A first degree of freedom is taken away by the data average. 
The other limiting factor of the number of components is the number of skull points. 
The numbers of coordinates of the known points are 65 ($N_0$), 220  ($N_{1}$) and 688 ($N_{2}$), which in the first case is less than the size of the learning base.
\begin{figure}[htbp]
\begin{center}
\includegraphics[width=15.20cm]{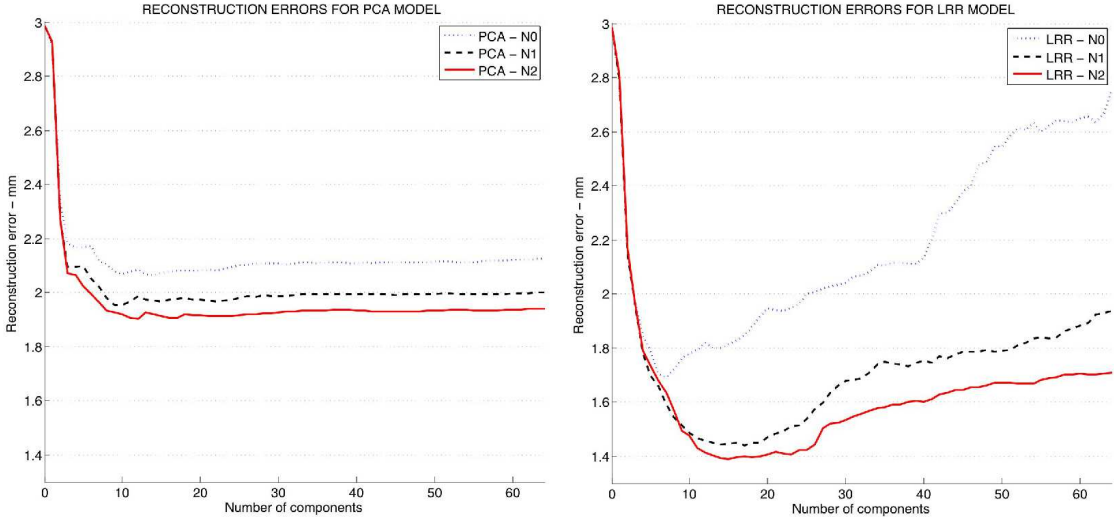}
\caption{Mean reconstruction errors}
\label{fig:Results}
\end{center}
\end{figure}

The reconstruction errors for the first 64 components are presented in Figure \ref{fig:Results} for the varying number of points and following the statistical methods presented in sections 3.2 and 3.3. 
For the first method, the reconstruction errors decrease greatly for the first parameters until a minimal value is achieved, then the reconstruction errors increase lightly. 
The minima of the mean reconstruction errors are 
$2.07 \pm 0.78$ mm ($N_0$), $1.95 \pm 0.73$ mm ($N_1$), $1.90 \pm 0.72$ mm ($N_2$)) and are achieved for 10 components ($N_0$,$N_1$) or 12 components. 
For LRR models, reconstruction errors follow another behaviour: reconstruction errors increase greatly after the optimal number of components. 
The minima of the mean reconstruction errors are $1.69 \pm 0.40$ mm achieved for 7 components ($N_0$),
$1.44 \pm 0.36$ mm achieved for 14 components ($N_1$) and $1.39 \pm 0.26$ mm achieved for 15 components ($N_2$). 
For all models, the increase in the number of skull points corresponds to a decrease in the reconstruction error. 
It also results in the increase in the number of selected components. 
The numbers of skull semi-landmarks are defined by the initial configuration of the mesh linking the anatomical landmarks. 
Improvements to the point correspondence procedure of the skull semi-landmarks will enable more points to be obtained as well as trying to find the optimal number of skull semi-landmarks. 
The minima of the reconstruction error for the LRR models are a half millimetre below the minima obtained for the PCA models. 
The selection of the number of components for the model has not the same consequences for the two methodologies. 
The reconstruction error for the maximum number of components in PCA models is near the minimal error whereas for LRR models the validation of the number of components must be carried out.
 
\begin{figure}[htbp]
\begin{center}
\includegraphics[width=12.0cm]{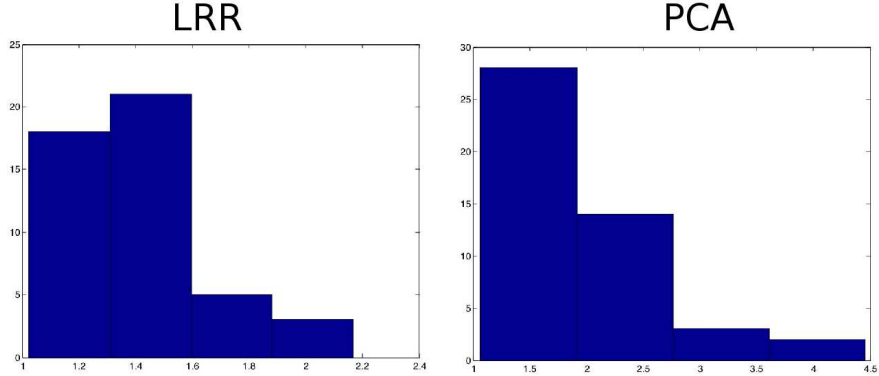}
\caption{histogram of the individual reconstruction errors for N2 points}
\label{fig:Hist}
\end{center}
\end{figure} 
 
\begin{figure}[h]
\begin{center}
\includegraphics[width=9.0cm]{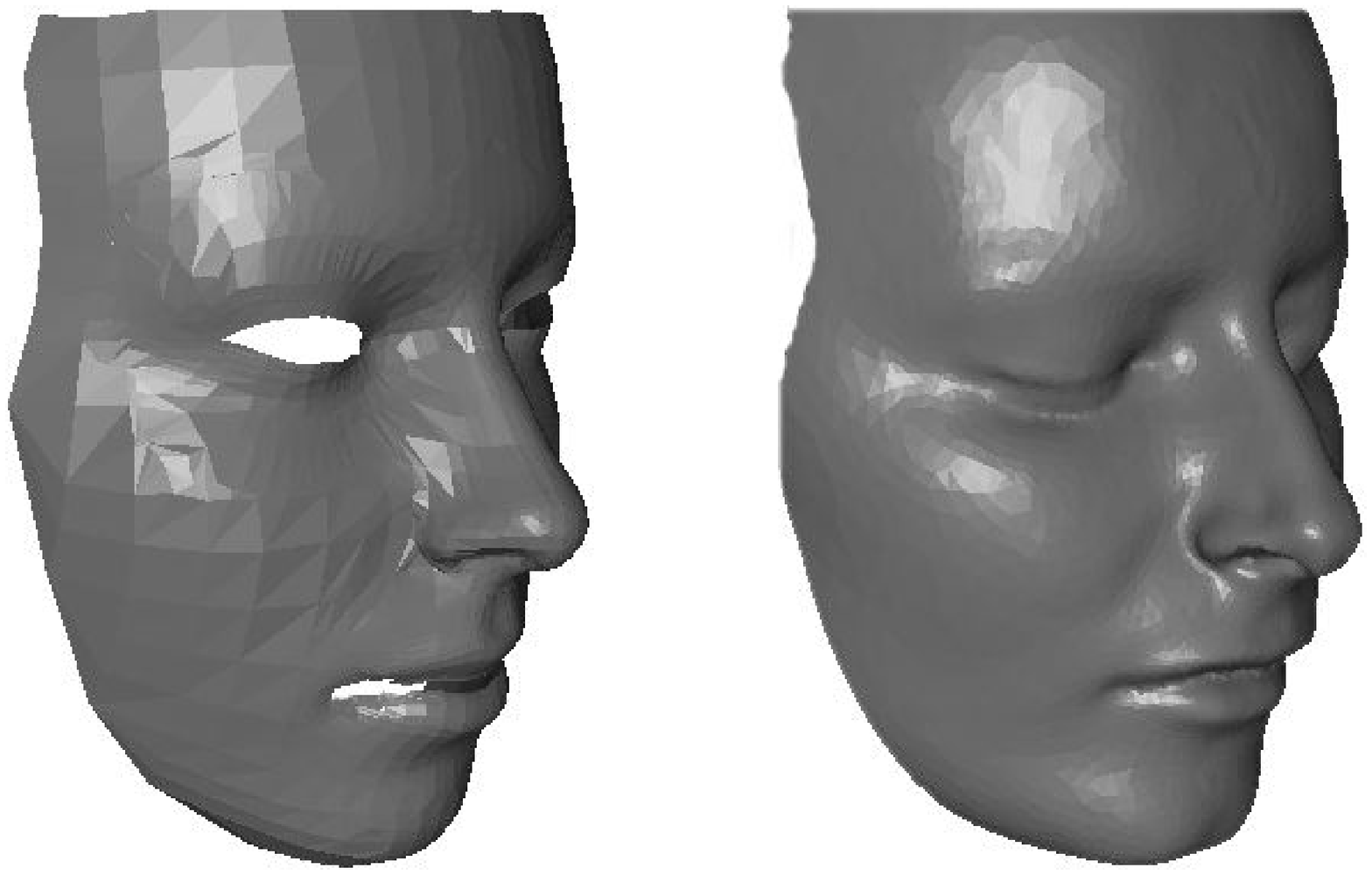}
\caption{Example of reconstructed complete face (left) and reference (right)}
\label{fig:Example1}
\includegraphics[width=15.0cm]{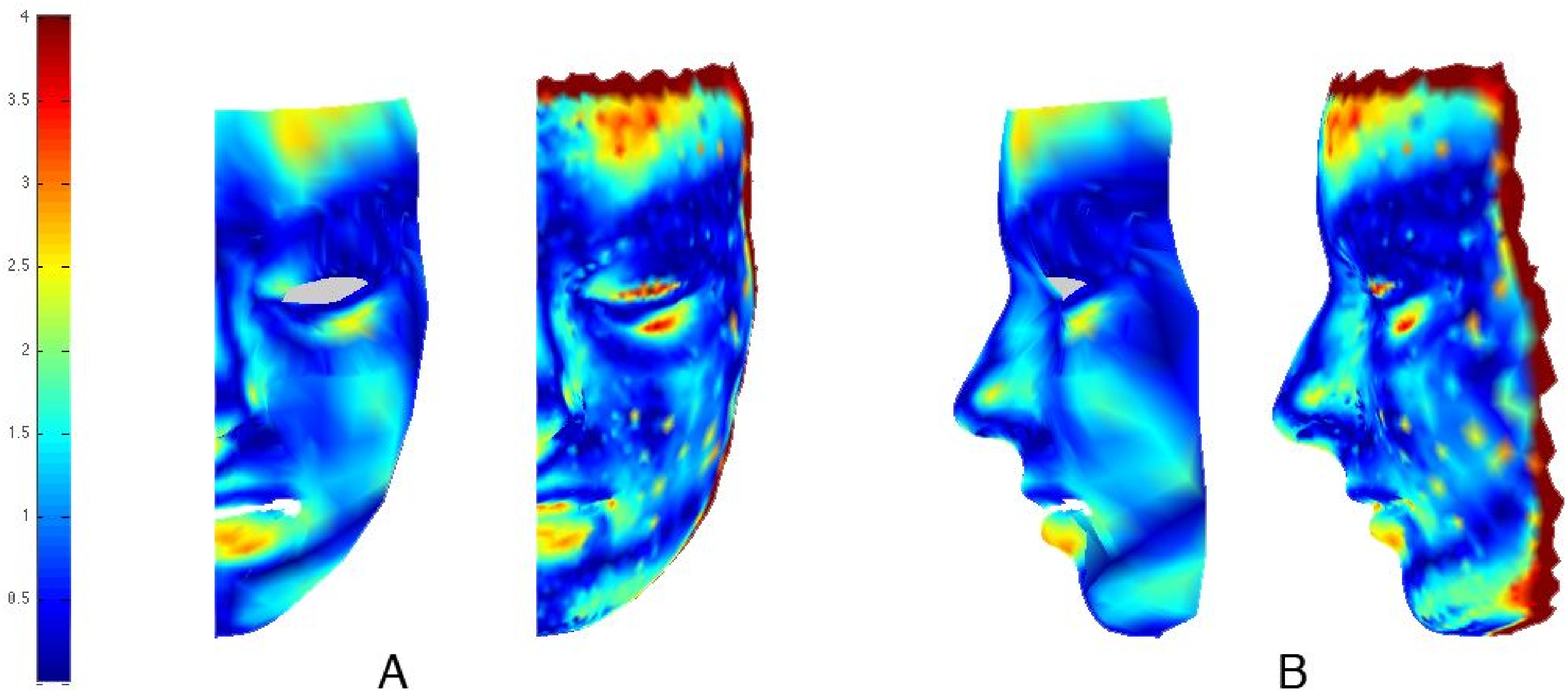}
\caption{Distance cards of a reconstruction: distribution of estimation errors over the surface
(A) full frontal: predicted (left) reference(right) 
(B) profile: predicted (left) reference (right). 
The color plot shows the errors from 0.0 mm (blue) to 4.0 mm (red).}
\label{fig:Example2}
\end{center}
\end{figure}

The model with the best results is the LRR model with $N_2$ points with 15 components. 
For this number of components, the individual errors range from 1.02 mm to 2.16 mm. 
Figure \ref{fig:Hist} (left) shows the histogram of the individual reconstruction error. 
For comparison purpose, the histogram of the reconstruction errors for the best PCA model is shown Figure  \ref{fig:Hist} (right). 
The individual errors range from 1.07 mm to 4.45 mm. 
An example of facial reconstruction is presented in Figure \ref{fig:Example1}. 
The resulting spatial maps of the quality of the reconstruction procedure can be seen in Figure \ref{fig:Example2}. 
The distance card from the predicted points toward the original surface has a mean error of 1.08 mm (left). 
For this case, the highest reconstruction errors are located on the superior part of the forehead and on the lips. 
This part of the forehead corresponds to a zone with no skull landmarks. 
The position of the lips is influenced by the mouth overture, where  variations occur during the acquisition. Similar remarks can be made concerning the more localised error in the regions of the nose and of the eyes.

\begin{figure}[htbp]
\begin{center}
\includegraphics[width=15.0cm]{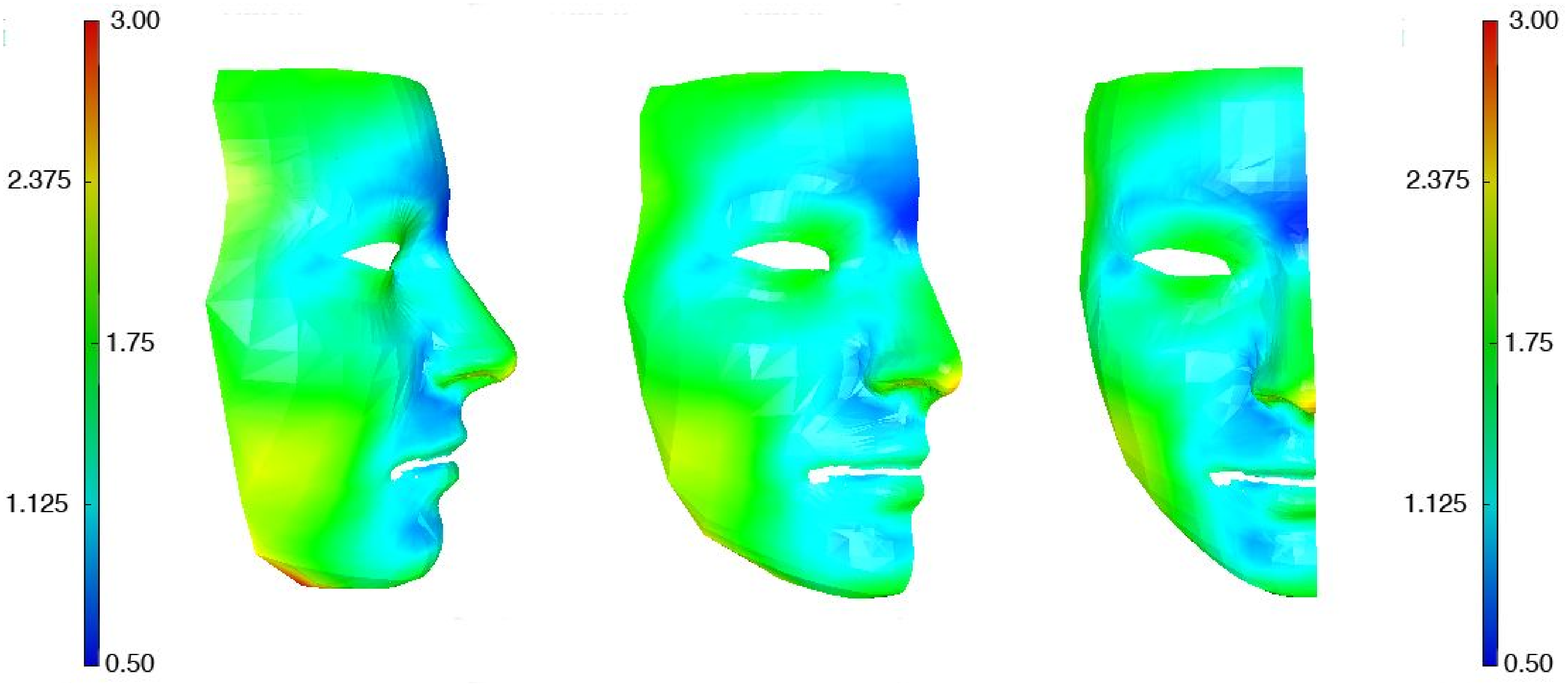}
\caption{Mean of the local error. The color plot shows the distribution over the surface from 0.0 mm (blue) to 3.0 mm (red).}
\label{fig:FaceMean}
\includegraphics[width=15.0cm]{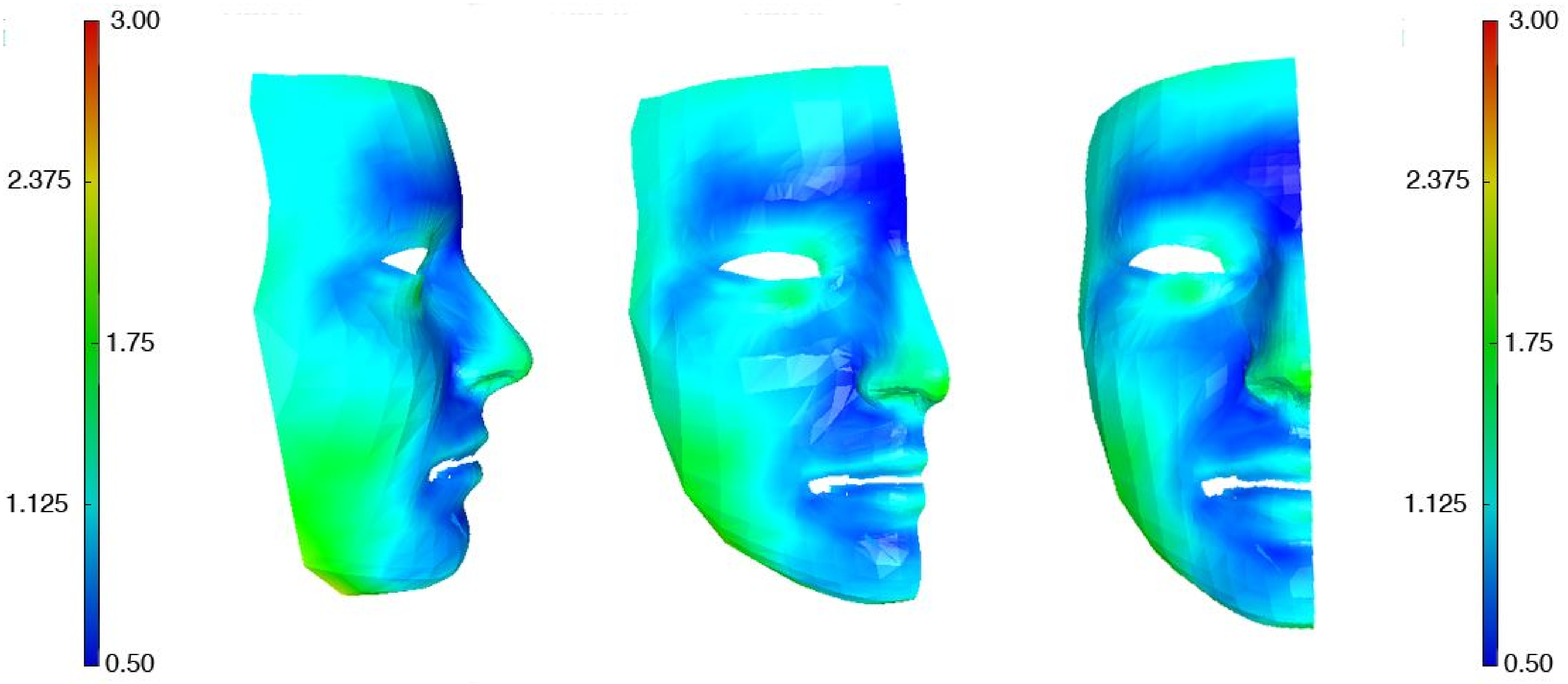}
\caption{Standard deviation of the local error.The color plot shows the distribution over the surface from 0.0 mm (blue) to 3.0 mm (red).}
\label{fig:FaceStd}
\includegraphics[width=15.0cm]{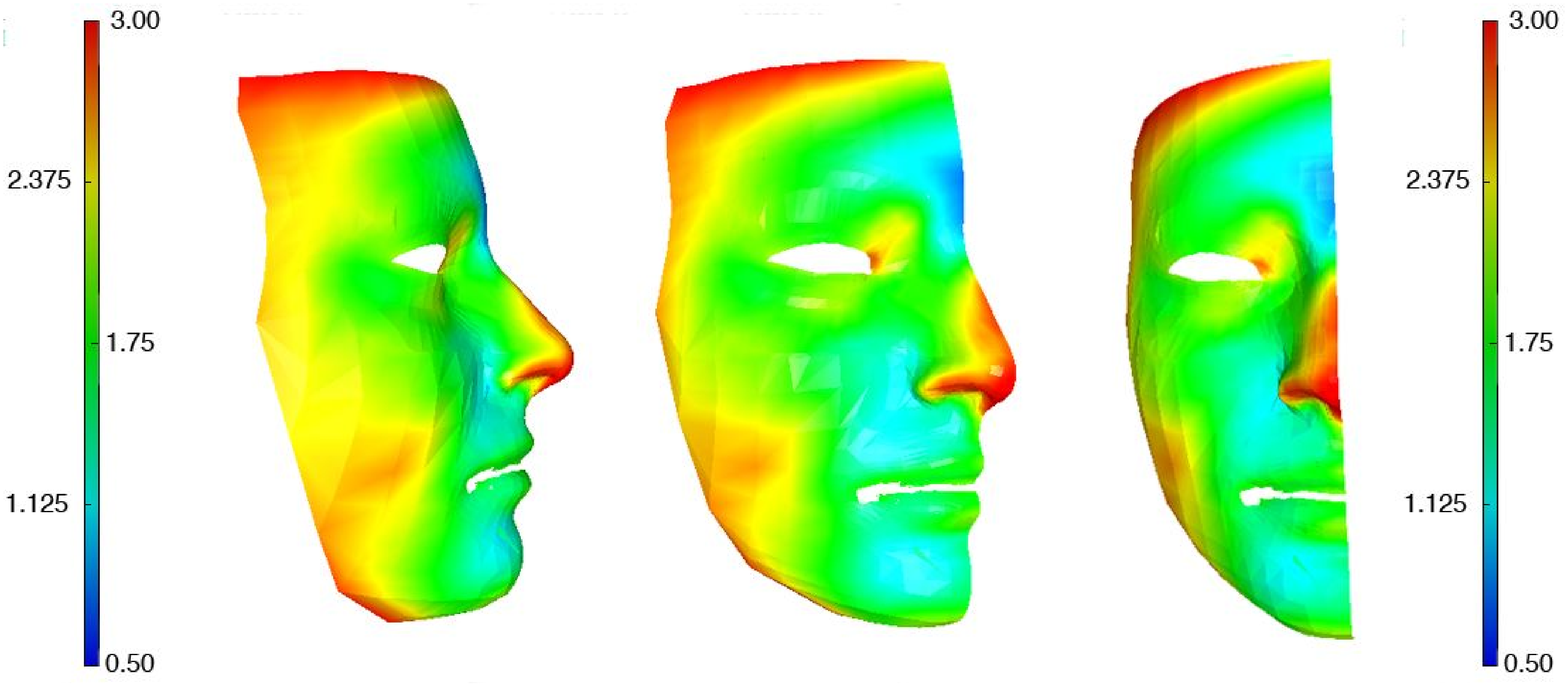}
\caption{Mean of the local error for PCA model. The color plot shows the distribution over the surface from 0.0 mm (blue) to 3.0 mm (red).}
\label{fig:FacePCA}
\end{center}
\end{figure}

For the optimal component number, we calculate mean and standard deviation for each predicted point of the mask. 
The mean of the point errors correspond to the reconstruction error (1.39 mm) and the standard deviation by points is then 1.02 mm. 
The resulting spatial maps of the quality of the reconstruction procedure for the optimal number of parameters can be seen in Figure \ref{fig:FaceMean} for the local mean and Figure \ref{fig:FaceStd} for the local standard deviation. 
The facial area with the highest reconstruction error (3 mm) corresponds to the boundary between the neck and the chin and is an artifact of the point correspondence step. There is no explicit correspondence to fix the limits of the mask in these zone. 
Others regions where the prediction error is high (2 mm to 3 mm) are the masseter region and the tip of the nose. 
These regions have few skull-landmarks and the bones do not support the soft-tissue for a large part of the cheeks. 
The regions with the smallest errors (0.5 mm to 1 mm) are concentrated towards the middle of the face, a part where the number of skull landmarks is high and where the inter-subject correspondence between the face meshes is more constrained. 
Figure \ref{fig:FacePCA} shows the distance map obtained for the best PCA model. The repartition of the error is similar with a greater amplitude.\\

\section{Discussion}

We compare our results to those of \cite{Claes2006} and \cite{Vandermeulen2006}. 
Among reconstruction techniques, the technique described in \cite{Claes2006} is close to the PCA model reconstruction techniques described here, with a supplementary deformation phase after the statistical prediction. 
The study is conducted on a database of 118 samples and use 57 landmarks for prediction. 
The number of face points obtained is likely to be high as the face meshes presented seem to tend towards surfaces. 
The reconstruction error corresponds roughly to our point-to-surface reconstruction errors. 
The mean reconstruction error observed is 1.14 mm with a standard deviation of 1.04 mm. 
The highest reconstruction errors (4 mm) are located in the chin and eye regions. 
Then, the region of the cheeks and the nose contains reconstruction errors of up to 2 mm, with a notable exception of the tip of the nose which is predicted separately. 
Regarding the smaller database and difference in the points correspondence step, our method achieves similar results with a generally simpler methodology. 
Concerning the distribution of the error on the face, we share similar zones in the cheeks and the nose whereas the chin and eye regions show smaller errors. The quality of prediction in the latter regions is more susceptible to be influenced by the point correspondence step of the soft-tissue surfaces.

The technique developed in \cite{Vandermeulen2006} is based on the use of continuous surface and the study conducted on 20 samples. The mean reconstruction error is 1.9 mm with a standard deviation of 1.7 mm. The  largest reconstruction errors (2-3 mm on average) occurs on the nostrils and masseter region. We surpass those results, however based on a smaller database. We can add that the regions with large reconstruction errors coincide once more.

A local method of facial reconstruction proposed by \cite{Tilotta2007} combines prediction obtained on surface patches, delimited by landmarks. The study has been performed on two regions: the nose region and the chin region. Local mean reconstruction errors of $1.40 \pm 0.25$ mm (nose) and $1.51 \pm 0.67$ mm (chin) occur for our method. The results presented in this report outperform these estimations with a mean reconstruction error of 0.99 mm, which motivates us to consider more local procedure in the reconstruction process.

\section{Conclusion}
We proposed a statistical method for 3D computerised forensic facial reconstruction, based upon Latent Root Regression. 
In conjunction, we have evaluated the influence of the number skull landmarks used. 
This method was based on an homogeneous database described in \cite{Tilotta2008arXiv}. 
As with most techniques based on a data analysis framework, the presented method relies on the use of a common set of points for the description of the individuals. 
In this set of points, anatomical skull landmarks are completed iteratively by points located upon geodesic curbs linking the anatomical landmarks. 
It enables us to artificially increase the number of skull landmarks. 
Facial landmarks are obtained using a mesh-matching algorithm between a common reference mesh and the individual soft-tissue surface meshes.

The regression problem of predicting the positions of the semi-landmarks of the soft-tissue surface knowing the positions of the skull landmarks has been resolved using two methodologies:  the building of a linear regression model, based on PCA and the use of Latent Root Regression. 
The accuracy of the reconstructions made by the methods was measured by leave-one-out cross-validation. 
First these results show the impact of an increased number of skull landmarks for the accuracy of the error computation of the reconstruction for both statistical methods. 
On this same set of couples of points, the LRR model shows better results whatever the number of components and the number of points. 
These results were discussed in regard to other facial reconstruction methods and performs comparatively to these other methods.

Some extensions can be proposed to the reconstruction method. 
First of all, having a larger database will increase the flexibility of the model. 
The more surface examples the model has, the better it learns the relationship between the two surfaces. 
Secondly, improvements of the correspondence steps will greatly benefit the reconstruction process. 
Improvement in the geodesics point correspondence method will help define an optimal number of skull landmarks. Improvements in the point correspondence step for the face will reduce the disparity observed in the computation of the distances between the reconstructed and the reference surface.
Lastly an automatic extraction of the anatomical landmarks from the skull would make the complete reconstruction pipeline automatic.

\end{document}